\documentclass[conference]{IEEEtran}
\IEEEoverridecommandlockouts
\usepackage{cite}
\usepackage{amsmath,amssymb,amsfonts}
\usepackage{algorithmic}
\usepackage{graphicx}
\usepackage{textcomp}
\usepackage{xcolor}
\usepackage{booktabs}
\usepackage{algorithm}
\usepackage{multirow}
\usepackage[normalem]{ulem}
\usepackage{hyperref}
\usepackage{authblk}
\usepackage{balance}
\def\BibTeX{{\rm B\kern-.05em{\sc i\kern-.025em b}\kern-.08em
    T\kern-.1667em\lower.7ex\hbox{E}\kern-.125emX}}

\begin{document}

\title{FR-DETR: Frequency and Recurrent Feature Refinement for Robust Object Detection \\ under Adverse Weather}

\author[1]{Tuan-Duc Nguyen}
\author[2,$\dagger$]{Duc-Trong Le\thanks{\textsuperscript{$\dagger$}Corresponding Author: trongld@vnu.edu.vn}}

\affil[1]{FPT Software AI Center}
\affil[2]{VNU University of Engineering and Technology}

\maketitle

\begin{abstract}
Object detection under adverse weather remains challenging due to severe visual degradations and domain shifts. Existing enhancer-based approaches attempt to improve detection by cascading an enhancer with a detector, but they introduce redundant feature extraction and incur high computational cost with limited accuracy gains when paired with SOTA detectors. 
We propose FR-DETR, a detector-centric framework that refines features rather than images, focusing enhancement on regions of interest and leveraging frequency-domain cues. Specifically, we design (I) a Frequency Refinement Module that dynamically separates and reweights low- and high-frequency components to improve foreground-background discrimination, and (II) a Recurrent Focus Refinement Module (RFRM) that iteratively refines features using coarse predictions as guidance. 
Extensive experiments demonstrate that FR-DETR achieves superior detection accuracy under adverse weather while being significantly more computationally efficient than enhancer-based methods. Our implementation is available at \href{https://github.com/ducnt1210/FR-DETR}{Link}.
\end{abstract}  

\begin{IEEEkeywords}
Object Detection, Adverse Weather, Efficiency
\end{IEEEkeywords}

\section{Introduction}
\label{sec:intro}

\begin{figure}[t]
\begin{center} 
   \includegraphics[width=0.95\linewidth,height=250px]{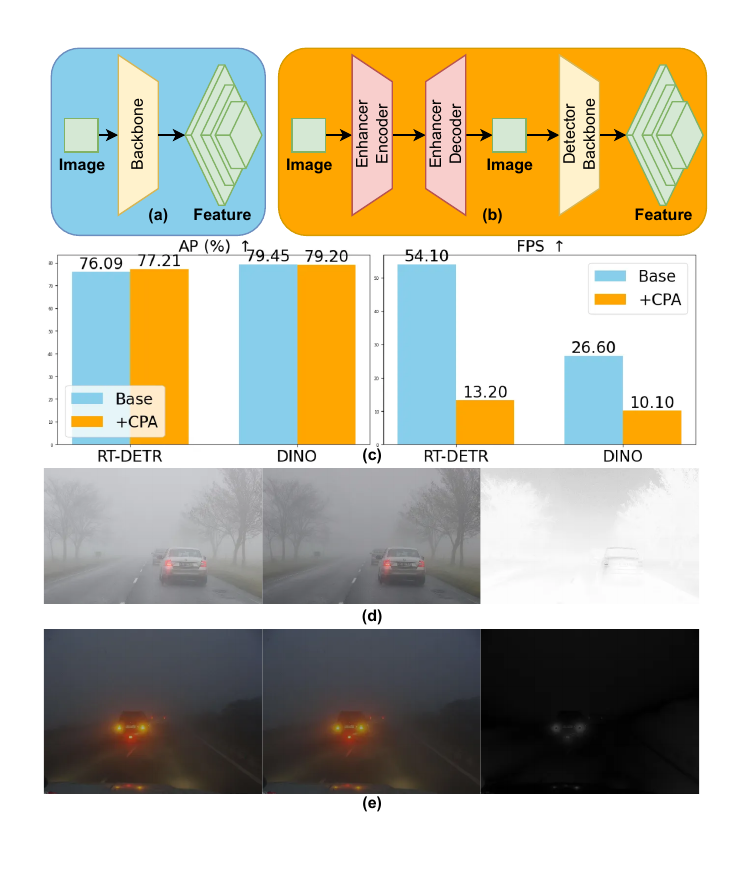}
\end{center}
\vspace*{-7mm}
   \caption{(a) Standard object detection pipeline, which extracts multi-scale features directly from the input image. (b) Enhancement-based approach, which performs redundant feature extraction, resulting in significant computational inefficiency. (c) Empirical evidence showing that current enhancer-based methods provide minimal performance gains while incurring substantial decrease in FPS. (d) and (e) from left to right are degraded images, enhanced images from an enhancer and their differences.}
\label{fig:motiv}
\end{figure}

Object detection underpins autonomous driving and surveillance, yet performance degrades sharply in adverse weather, e.g., fog, rain, and snow, due to severe distortions and structured noise that induce a domain shift from clean training data. Most detectors are trained on favorable conditions; data augmentation~\cite{Gupta_2024_WACV} is simple but fails under heavy noise/occlusion. Pre-restoration~\cite{histoformer,dehaze_rel,Hu_2019_CVPR} can look good but often harms detection~\cite{Gupta_2024_WACV}. Multitask schemes~\cite{MAET} need paired data and are hard to train~\cite{reforde}. End-to-end methods~\cite{iayolo,gdip,cpa} improve robustness but still face several limitations. First, as in Fig.~\ref{fig:motiv}(a,b), current pipelines duplicate feature extraction: an enhancer reconstructs an image that the detector backbone re-encodes, inflating compute and hurting real-time use. Empirically (Fig.~\ref{fig:motiv}(c)), CPA-Enhancer~\cite{cpa}+RT-DETR~\cite{RTDETR} or DINO~\cite{dino} slashes FPS with marginal or negative accuracy gains; trends are similar across enhancers. Second, enhancements are misaligned with detection: Fig.~\ref{fig:motiv}(d,e) shows global, redundant and non-selective edits over boundaries and distant vehicles.
Adverse weather often causes foreground objects to blend into the background, a phenomenon that also characterizes camouflage object detection. Recent works in camouflage detection~\cite{Freq_camo_CVPR} demonstrate that explicitly modeling frequency cues helps disentangle object structures from background clutter. More details about related works are shown in \textbf{Appendix}.

Motivated by these issues, we migrate enhancement from the image domain into the feature space of the detector, thereby eliminating redundant image-level feature extraction. We propose \textbf{F}requency and \textbf{R}ecurrent feature refinement \textbf{DE}tection \textbf{TR}ansformer framework (FR-DETR) for object detection under adverse weather. It consists of two primary components: the Recurrent Focus Refinement Module (RFRM) and the Frequency Refinement Module (FRM). 
The former leverages coarse predictions from the previous stage as guidance to refine features iteratively, concentrating computation on regions of interest. This design steers enhancement toward detection-relevant structures while improving computational efficiency under adverse weather conditions. The latter leverages frequency-domain information to strengthen foreground--background separability under adverse conditions. Specifically, it operates in feature space, dynamically decoupling low- and high-frequency components to extract complementary cues.

Our main contributions are as follows:
\begin{itemize}
    \item We provide an empirical analysis demonstrating that existing enhancer-based approaches are computationally inefficient and poorly suited for integration with state-of-the-art transformer-based detectors.
    \item We propose a Frequency Refinement Module that dynamically separates low- and high-frequency components in feature space, extracting complementary cues to improve foreground--background discrimination under adverse weather.
    \item We introduce a Recurrent Focus Refinement Module that iteratively refines features using coarse predictions as guidance, directing enhancement toward regions of interest most relevant for detection.
    \item Extensive experiments on multiple benchmarks show that our FR-DETR achieves superior accuracy compared to enhancer-based methods while being significantly more computationally efficient.
\end{itemize}

\section{Methodology}

\begin{figure}[t]
\begin{center} 
   \includegraphics[width=1.0\linewidth]{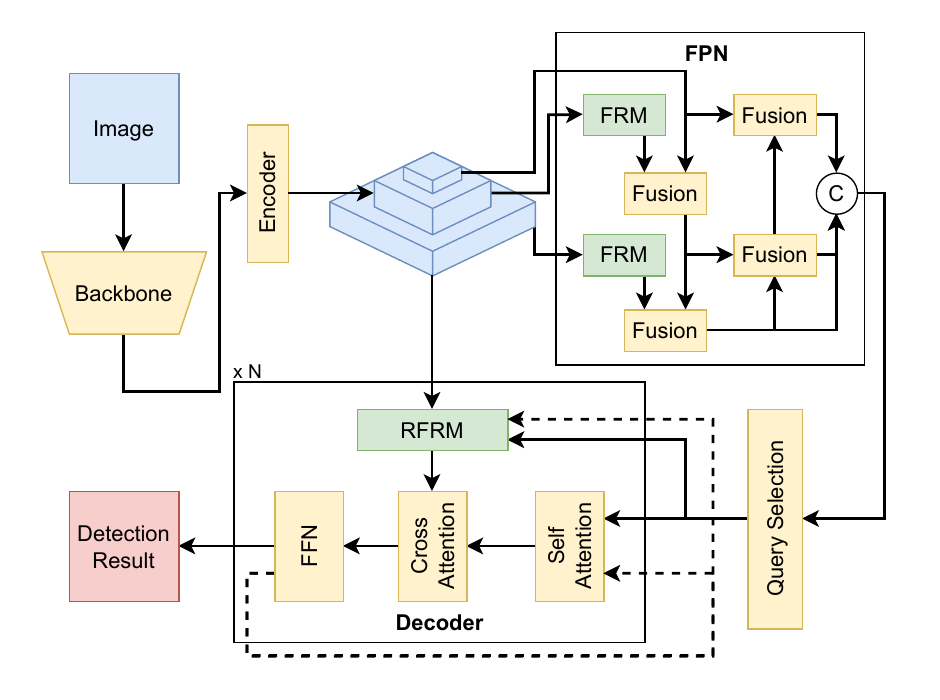}
\end{center}
\vspace*{-2mm}
   \caption{Overview of the FR-DETR architecture. The core RT-DETR pipeline, including the backbone, encoder, FPN, and multi-stage decoder (highlighted in yellow), is preserved. We introduce two modules (green): (I) a Frequency Refinement Module (FRM) inserted into the FPN pathway, and (II) a Recurrent Focus Refinement Module (RFRM) that operates across decoder stages.}
\label{fig:Overview}
\end{figure}
Our method builds upon RT-DETR \cite{RTDETR}, which employs a backbone for multi-scale feature extraction, followed by an encoder that processes the deepest feature map. A Feature Pyramid Network (FPN) then facilitates cross-scale feature interaction. Subsequently, the top-k image features are selected as initial object queries for the decoder, which iteratively refines predictions across multiple stages to produce object categories and bounding boxes. To adapt this framework for adverse weather conditions, we introduce two key architectural enhancements: (I) a Frequency Refinement Module (FRM) that leverages frequency-domain information to provide robust cues in weather-degraded scenes, and (II) a Recurrent Focus Refinement Module (RFRM) that progressively enhances features in regions of interest across decoder stages. The overall architecture of FR-DETR is illustrated in Fig. \ref{fig:Overview}.

\begin{figure}[t]
\begin{center} 
   \includegraphics[width=1.0\linewidth]{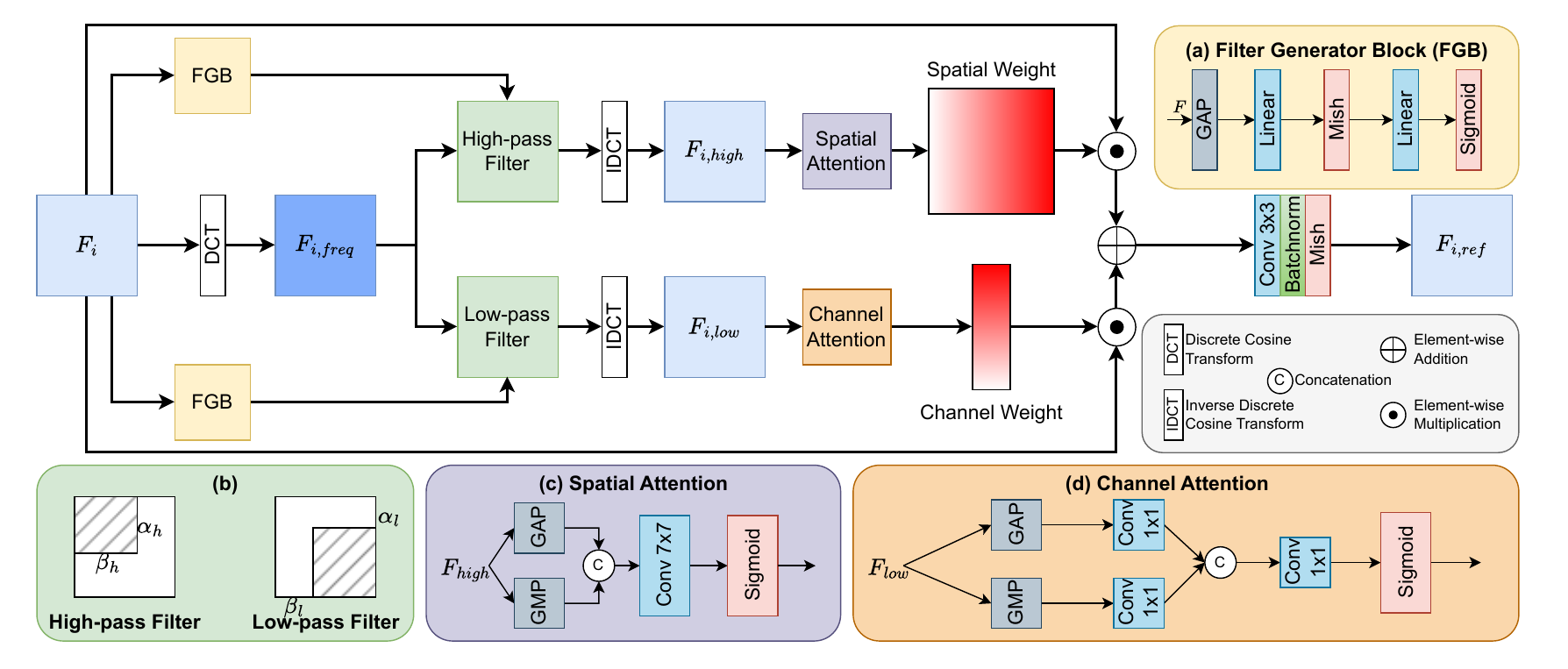}
\end{center}
\vspace*{-5mm}
   \caption{The architecture of Frequency Refinement Module (FRM). Given an input feature map, a Filter Generator Block predicts a pair of high-pass and low-pass frequency filters. The feature is transformed to the frequency domain via DCT, filtered by the predicted frequency masks, and mapped back by inverse DCT. GAP stands for global average pooling operation while GMP is global max pooling. The filtered high and low frequency responses are fused with the original feature through spatial and channel attention. An output projection then produces the refined feature, enriching the representation with complementary cues from the frequency domain.}
\label{fig:FRM}
\end{figure}

\subsection{Frequency Refinement Module}
We propose the Frequency Refinement Module (FRM) that dynamically extracts, reweighs, and fuses high/low-frequency responses to refine detector features. Fig.~\ref{fig:FRM} illustrates the architecture of FRM.
Specifically, we insert the FRM at the lateral connections of the FPN, replacing the standard skip fusion between the upsampled top-down feature and the corresponding lateral feature. Low-level features are rich in fine spatial detail but lack semantic abstraction, whereas high-level features are semantically strong yet spatially coarse due to upsampling. FRM counteracts this misalignment by extracting informative high- and low-frequency responses from the lateral feature and integrating them with the top-down feature.

Given the feature map $F_i \in \mathbb{R}^{C_i \times H_i \times W_i}$, we first compute its frequency representation $F_i^{\mathrm{freq}} \in \mathbb{R}^{C_i \times H_i \times W_i}$ via 2D DCT. Since standard DCT lacks explicit frequency separation, we design a lightweight Filter Generator Block (FGB, illustrated in Fig.\ \ref{fig:FRM}(a)) that predicts two data-dependent frequency masks (high-pass and low-pass), which act as boundaries in the frequency domain.
The process of FGB can be formally expressed as:
\begin{equation}
(\alpha_{f},\beta_{f}) = \delta(\text{Linear}(\phi(\text{Linear}(\text{GAP}_s(F_i)))))
\end{equation}
where $f \in \{high, low\}$ denoting high- and low-frequency bands; $\alpha_{f},\beta_{f} \in [0,1]$ control the height and width cutoffs. $\text{GAP}_s$ denotes spatial global average pooling, $\phi$ is the Mish activation and $\delta$ indicates the sigmoid function.
The hard mask $M^{\text{hard}}_f \in \{0,1\}^{H_i \times W_i}$ for low and high-pass are as illustrated in Fig.~\ref{fig:FRM}(b) where dashed regions equal 0 while the remaining regions are 1. Therefore, the high-pass mask attenuates coefficients around the DC corner (upper-left), while the low-pass mask suppresses high-frequency components (bottom-right).

However, naively using hard thresholding yields non-differentiable masks. To enable end-to-end learning, we adopt the Straight-Through Estimator (STE)~\cite{STE}. We formulate a differentiable soft mask $M^{\text{soft}}_f \in [0,1]^{H_i \times W_i}$ (detailed in \textbf{Appendix}) to approximate the gradient. Formally, the STE combination is as:
\begin{align}
M_f = \mathrm{sg}(M_f^{\text{hard}} - M_f^{\text{soft}}) + M_f^{\text{soft}},
\end{align}
where $\mathrm{sg}(\cdot)$ denotes the stop-gradient operator.
This ensures the forward pass utilizes the crisp binary mask $M^{\text{hard}}_f$, while gradients propagate through the smooth $M^{\text{soft}}_f$.

After generating single-channel $M_{high}$ and $M_{low}$ filters, we conduct masking to every channel of $F_{i,freq}$ and use inverse DCT to create high- and low-frequency features $F_{high}$ and $F_{low}$. Finally, we use different attention strategies \cite{cbam} to extract informative cues from the frequency domain.

\noindent \textbf{Spatial Attention Path.}
High-frequency features contain rich boundary cues and fine-grained details that are critical for accurate localization, especially under adverse weather. 
We therefore employ a spatial attention branch to inject this information into the feature representation. 
Given the high-frequency response $F_{\mathrm{high}}$, we compute channel-wise average and max pooling projections
before concatenating them along the channel dimension. Subsequently, we apply a large-kernel convolution followed by a sigmoid to obtain the spatial attention $A_s \in [0,1]^{H_i \times W_i}$ (see Fig.~\ref{fig:FRM}c).

\noindent \textbf{Channel Attention Path.}
Low-frequency components capture global structure and semantic context, which are essential for robust object classification under adverse weather, where significant information may be degraded or missing. We employ a channel attention branch to reintegrate these cues into the feature representation. This mechanism assigns adaptive weights to each channel, emphasizing informative responses while suppressing less relevant ones. Specifically, given the low-frequency feature $F_{\mathrm{low}}$, we first apply GAP and GMP along the spatial dimensions, producing two descriptors of size $\mathbb{R}^{C_i}$. Each descriptor is passed through a $1{\times}1$ convolution, and the outputs are concatenated along the channel dimension, followed by a final $1{\times}1$ projection and a sigmoid activation to obtain the channel attention weight $A_c \in [0,1]^{C_i}$ (see Fig.~\ref{fig:FRM}d).

After extracting complementary cues from the high and low-frequency branches, we fuse them with element-wise summation followed by a projection as follows: 
\begin{equation}
    F_{i,ref} = \phi(\mathrm{BN}(\mathrm{Conv}_{3\times3}(F_i\odot(A_c + A_s))))
\end{equation}
where $F_{i,ref} \in \mathbb{R}^{C_i \times H_i \times W_i} $ is the frequency refined feature, $\phi$ denotes the Mish activation function and $\mathrm{BN}$ is batch norm. This operation enriches the representation with boundary-preserving and semantically coherent cues, yielding features that are better suited for detection in adverse weather.

\subsection{Recurrent Focus Refinement Module}
\begin{figure}[t]
\begin{center} 
   \includegraphics[width=1.0\linewidth]{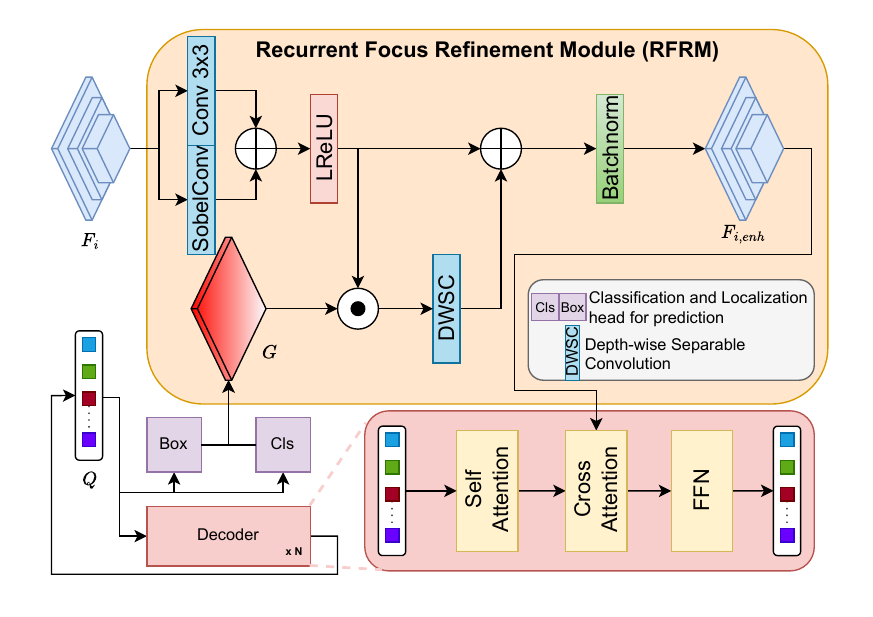}
\end{center}
\vspace*{-5mm}
   \caption{The architecture of Recurrent Focus Refinement Module (RFRM). At each decoder stage, object queries $Q$ produce preliminary box and class predictions, which are used to generate a guidance map $G$. This map highlights object-centric regions and is fused with the image feature to selectively enhance these regions. The refined feature is then fed into the cross-attention block of the next decoding stage, enabling progressive, region-focused refinement across stages.}
\label{fig:RFRM}
\end{figure}

\begin{table*}[ht]
\centering
\caption{Results under foggy conditions using 5-class models. Best and second-best scores are bold and underlined.}
\renewcommand{\arraystretch}{1.0}
\begin{tabular}{@{}llcccccccccccc@{}}
\toprule
\multirow{2}{*}{Approach} & \multirow{2}{*}{Model} & \multicolumn{3}{c}{V\_n\_ts} & \multicolumn{3}{c}{V\_f\_ts} & \multicolumn{3}{c}{RTTS} & \multicolumn{3}{c}{Performance} \\ 
\cmidrule(lr){3-5} \cmidrule(lr){6-8} \cmidrule(lr){9-11} \cmidrule(l){12-14}
 & & AP & AP$_{50}$ & AP$_{75}$ & AP & AP$_{50}$ & AP$_{75}$ & AP & AP$_{50}$ & AP$_{75}$ & Params(M) & GFlops & FPS \\ 
\midrule
Direct & RT-DETR \cite{RTDETR} & 76.90 & 81.66 & 64.85 & 76.09 & 80.93 & 63.97 & 41.74 & 46.97 & 26.96 & 20.09 & 61.12  & 54.1 \\
\midrule
\multirow{6}{*}{Preprocessing} & RT-DETR \cite{RTDETR} & \_ & \_ & \_ & 73.24 & 77.36 & 62.14 & 41.60 & 46.23 & 28.51 & 20.09 & 61.12  & 54.1 \\
& ConvIR \cite{ConvIR} & \_ & \_ & \_ & 74.97 & 79.20 & 63.78 & 41.67 & 46.37 & 28.21 & 25.62  & 546.27 & 16.8 \\
& PHATNet \cite{PhatNet} & \_ & \_ & \_ & 73.17 & 76.98 & 62.40 & 38.36 & 42.49 & 26.31 & 23.83 & 403.11 & 24.9 \\
& Histoformer \cite{histoformer} & \_ & \_ & \_ & 75.20 & 79.64 & 63.85 & 41.55 & 46.15 & \underline{28.71} & 36.71 & 1075.12 & 1.5 \\
& DCPT \cite{dcpt} & \_ & \_ & \_ & 76.61 & 80.93 & 65.23 & 39.72 & 44.22 & 26.84 & 87.98 & 971.74 & 12.1 \\
& MoceIR \cite{moceir} & \_ & \_ & \_ & 75.76 & 79.84 & 64.45 & 39.09 & 43.36 & 26.60 & 31.57 & 563.37 & 2.1 \\
\midrule
\multirow{3}{*}{Union} & MGDIP \cite{gdip}   & 77.34 & 81.90 & 65.67 & \underline{77.45} & \underline{82.14} & \underline{66.14} & 41.68 & 46.30 & 27.81 & 26.90 & 116.45 & 8.1  \\
& DENet \cite{denet}   & 77.50 & 82.10 & 65.74 & 76.39 & 81.22 & 64.24 & 41.80 & 46.97 & 27.13 & 20.13 & 65.33  & 45.0 \\
& CPA \cite{cpa}     & \underline{77.61} & \underline{82.43} & \underline{65.89} & 77.21 & 81.94 & 65.55 & \underline{42.01} & \underline{47.24} & 27.55 & 23.94 & 217.91 & 13.2 \\
\midrule
Our & \textbf{FR-DETR} & \textbf{77.99} & \textbf{82.58} & \textbf{66.66} & \textbf{77.86} & \textbf{82.41} & \textbf{66.36} & \textbf{43.45} & \textbf{48.41} & \textbf{29.41} & 23.28 & 96.04  & 31.9 \\ 
\bottomrule
\end{tabular}
\label{tab:rtdetr_base}
\end{table*}

We replace image-level enhancement with a feature-space alternative. Our Recurrent Focus Refinement Module (RFRM) operates at each RT-DETR decoder stage, using coarse predictions from the previous stage to derive a focus signal that selectively refines features at stage $t$, without reconstruction. By confining enhancement to task-relevant regions directly in feature space, RFRM (I) eliminates redundant double encoding and (II) biases refinement toward object-centric cues robust to weather-induced degradations. An overview of the RFRM architecture is shown in Fig.~\ref{fig:RFRM}.

\noindent \textbf{Guidance map construction.}
At decoder stage $t$, object queries $Q_t \in \mathbb{R}^{N \times C}$ yield predictions ${(\hat{b}^{t}_n, \hat{p}^{t}_n)}$, where $\hat{b}^{t}_n$ are normalized boxes and $\hat{p}^{t}_n$ are class probabilities. To extract robust spatial priors, we must filter reliable predictions. We observe that hard confidence thresholding leads to a cold-start collapse: in early epochs, low confidence scores result in zero-mask guidance, destabilizing training. To mitigate this, we adopt a top-$k$ selection, which ensures a consistent guidance signal and stabilizes training.

For each of the selected queries, we choose confidence scores as the max confidence scores among all classes, then place an anisotropic Gaussian around each box center to encode proximity. We aggregate and normalize these responses into a guidance map $G \in [0,1]^{H_t \times W_t}$. Overlapping boxes naturally increase local density via summation, strengthening signals in crowded regions. The guidance map $G$ is then fused with the decoder feature to produce a region-focused refinement.

\noindent \textbf{Recurrent Focus Refinement Modules.}
After constructing the guidance map $G$, we incorporate it into RFRM to refine features at each decoder stage (Fig.~\ref{fig:RFRM}). For efficiency, we apply RFRM only to one out of three image features, which offers the best compute-accuracy trade-off; we will ablate this choice in our experiments. Given the input feature $F_i$, we use a two-branch projection to extract complementary cues:
a semantic branch to encourage channel interaction, and a visual branch to emphasize fine details. The branches are defined as:
\begin{equation}
    \begin{gathered}
        F_{\mathrm{sem}} = \mathrm{Conv}_{3\times 3}(F_i), \quad
        F_{\mathrm{vis}} = \mathrm{Sobel}_x(F_i) + \mathrm{Sobel}_y(F_i) \\
        F_{\mathrm{comb}} = \text{LReLU}\!\big(F_{\mathrm{sem}} + F_{\mathrm{vis}}\big)
    \end{gathered}
\end{equation}
where $\mathrm{Sobel}_x$ and $\mathrm{Sobel}_y$ denote depth-wise $3{\times}3$ convolutions whose kernels are initialized with the standard horizontal and vertical Sobel filters, respectively.

Given the combined feature $F_{\mathrm{comb}} \in \mathbb{R}^{C \times H \times W}$ and the guidance map 
$G \in [0,1]^{1 \times H \times W}$ broadcasted along channels, 
we apply element-wise modulation to get $F_{\mathrm{focus}}$.
To refine the focused representation with minimal overhead, we employ a depthwise separable convolution (DWSC) and a residual connection, followed by batch normalization:
\begin{equation}
    F_{\mathrm{enh}} \;=\; 
    \mathrm{BN}\!\Big(\mathrm{DWSC}(F_{\mathrm{focus}}) + F_{\mathrm{comb}}\Big),
\end{equation}
Here, the residual path preserves identity information, while DWSC enhances object-aware regions indicated by $G$. The refined feature $F_{\mathrm{enh}}$ is then fed to the cross-attention block in the decoder to improve representation in adverse weather.

\section{Experiments}
\label{sec:experiments}

\subsection{Datasets}

\noindent \textbf{Foggy Conditions.}
Consistent with prior work~\cite{cpa,iayolo,gdip}, we evaluate five classes (\emph{person, car, bus, bicycle, motorcycle}). We synthesize fog on Pascal VOC~\cite{pascal-voc-2007} via the atmospheric scattering model and utilize its test set (denoting the train set as V\_n\_tv and the test set as V\_f\_ts) and the real-world RTTS benchmark~\cite{rtts} for evaluation.

\noindent \textbf{Multiple Weather Conditions.}
To address the limitations of existing datasets regarding low intensity and single degradations, we adopt the Adverse Weathers dataset based on multiple open-access datasets, following \cite{MODE}. This benchmark includes high-intensity fog, rain, and snow, comprising 15,000 train and 3,000 evaluation images across six classes.

\noindent More datasets details are provided in the \textbf{Appendix}.

\subsection{Experimental Setup}

\noindent \textbf{Baselines.}
Our primary baseline is RT-DETR~\cite{RTDETR}. We further compare against prior enhancer-based pipelines, CPA, MGDIP, DENet~\cite{cpa,gdip,denet}, all trained jointly with RT-DETR for a fair comparison.

\noindent \textbf{Evaluation Strategy and Metrics.}
We evaluate against three paradigms: (I) Direct: Training detectors solely on degraded images; (II) Preprocessing: Using a frozen pretrained restorer before detection; and (III) Union: End-to-end training of enhancer and detector. Our FR-DETR is directly trained on degraded images like RT-DETR.
Performance is measured via COCO metrics reporting: $\text{AP}$, $\text{AP}_{50}$, and $\text{AP}_{75}$.

\noindent \textbf{Implementation Details.}
Experiments are conducted in PyTorch, where images are resized to $448^2$ with a batch size of 6. We use a pretrained ResNet-18VD backbone, optimized via AdamW with an lr of $10^{-4}$, weight decay of $10^{-4}$, and a LinearLR scheduler. Considering dataset scales and difficulties, the training epochs for foggy and multi-weather conditions are 48 and 72, respectively. Experiments use a single NVIDIA A100 (40GB), except CPA, which requires an A100 (80GB) due to memory constraints.

\subsection{Main Results} 
\noindent \textbf{RT-DETR with Enhancers on Foggy Condition}
Table~\ref{tab:rtdetr_base} summarizes the results across synthetic and real-world foggy benchmarks.
FR-DETR consistently outperforms all baselines across all metrics. Notably, on the real-world RTTS dataset, our method surpasses the second-best approach by \textbf{+1.44} $\text{AP}$, demonstrating superior performance without relying on explicit image-level enhancement.
We further evaluate the trade-off between accuracy and computational cost (Params, GFLOPs, FPS), measured on a single NVIDIA A100 at $640 \times 640$. 
FR-DETR achieves a superior accuracy-efficiency balance, maintaining high throughput comparable to the base detector.
In stark contrast, enhancer-based cascades suffer from significant overhead: MGDIP drops to unviable speeds (8 FPS), while CPA triples the computational cost to 218 GFLOPs.
This bottleneck largely stems from redundant feature extraction and expensive pixel-level reconstruction in the enhancer prior to re-encoding by the detector's backbone, which is avoided by our detector-centric design.

\begin{table}[t]
\centering
\caption{Additional results under adverse weathers using 6-class models. Best and second-best scores are bold and underlined.}
\renewcommand{\arraystretch}{1.0}
\begin{tabular}{@{}llccc@{}}
\toprule
\multirow{2}{*}{Approach} & \multirow{2}{*}{Model} & \multicolumn{3}{c}{Adverse Weathers} \\ 
\cmidrule(lr){3-5}
 & & AP & AP$_{50}$ & AP$_{75}$ \\ 
\midrule
Direct & RT-DETR \cite{RTDETR} & 8.22& 19.31& \underline{6.40}\\
\midrule
\multirow{4}{*}{Preprocessing} & RT-DETR \cite{RTDETR} & 3.78& 9.28& 2.65\\
& Histoformer \cite{histoformer} & 4.52& 10.65& 3.37\\
& DCPT \cite{dcpt} & 4.11& 9.84& 2.94\\
& MoceIR \cite{moceir} & 4.17& 9.94& 3.05\\
\midrule
\multirow{1}{*}{Union} & DENet \cite{denet}   & \underline{8.23} & \underline{19.45} & 6.34\\
\midrule
Our & \textbf{FR-DETR} & \textbf{10.82}& \textbf{23.39}& \textbf{9.09}\\ 
\bottomrule
\end{tabular}
\label{tab:rtdetr_aw}
\end{table}

\noindent \textbf{Performance on Adverse Weather.}
Table~\ref{tab:rtdetr_aw} presents the results on the challenging multi-weather benchmark. This dataset exhibits significantly higher complexity than the single-condition foggy benchmarks.
FR-DETR demonstrates robust generalization, outperforming the RT-DETR baseline by a substantial margin, notably improving $\text{AP}_{50}$ by \textbf{+4.08} points. Notably, this performance gap is much wider than that observed on the foggy dataset, suggesting that FR-DETR is advantageous in scenarios with severe and diverse degradations.

Please refer to the \textbf{Appendix} for additional results, ablations, detailed breakdowns of specific weather degradation results as well as feature maps and detection visualizations.

\subsection{Ablation Studies}

\begin{table}[t]
\centering
\caption{Ablation studies.}
\resizebox{0.48\textwidth}{!}{
\renewcommand{\arraystretch}{1.0}
\begin{tabular}{cccccc}
\toprule
& & & $\text{AP}$ & $\text{AP}_{50}$ & $\text{AP}_{75}$\\
\midrule
\multicolumn{6}{l}{(a) Importance of Main Components} \\
\hline
\textbf{Baseline} & \textbf{FRM} & \textbf{RFRM} & & &\\
\hline
\checkmark  & & & 76.09 & 80.93 & 63.97 \\
\checkmark & \checkmark & & 76.99 & 81.93 & 64.85 \\
\checkmark & & \checkmark & 76.85 & 81.59 & 65.19 \\
\checkmark & \checkmark & \checkmark & \textbf{77.86} & \textbf{82.41} & \textbf{66.36} \\
\hline
\multicolumn{6}{l}{(b) Impact of Frequency Cues} \\
\hline
\textbf{Baseline} & \textbf{High-Freq} & \textbf{Low-Freq} & & &\\
\hline
\checkmark  & & & 76.09 & 80.93 & 63.97 \\
\checkmark & \checkmark & & 76.63 & 81.63 & 64.66 \\
\checkmark & & \checkmark & 76.60 & 81.54 & 64.39 \\
\checkmark & \checkmark & \checkmark & \textbf{76.99} & \textbf{81.93} & \textbf{64.85} \\
\hline
\multicolumn{6}{l}{(c) Importance of the Filter Generator Block} \\
\hline
\multicolumn{3}{c}{\textbf{Filter Threshold}} & & &\\
\hline
\multicolumn{3}{c}{0.25} & 76.79 & 81.53 & 64.77 \\
\multicolumn{3}{c}{0.50} & 76.72& 81.74 & 64.48 \\
\multicolumn{3}{c}{0.75} & 76.73 & 81.73 & 64.73 \\
\multicolumn{3}{c}{1.00} & 76.54 & 81.53 & 64.21 \\
\multicolumn{3}{c}{Dynamic Filter \textbf{(Ours)}} & \textbf{76.99} & \textbf{81.93} & \textbf{64.85} \\
\hline
\multicolumn{6}{l}{(d) Importance of Guidance Map} \\
\hline
\multicolumn{3}{c}{Without Guidance Map} & 77.30 & 82.02 & 65.52 \\
\multicolumn{3}{c}{With Guidance Map \textbf{(Ours)}} & \textbf{77.86} & \textbf{82.41} & \textbf{66.36} \\
\hline
\multicolumn{6}{l}{(e) Selection of Feature Map Level} \\
\hline
\textbf{High Res} & \textbf{Mid Res} & \textbf{Low Res} & & &\\
\hline
\checkmark  & & & \textbf{77.86} & 82.41 & \textbf{66.36} \\
& \checkmark & & 77.69 & \textbf{82.48} & 66.00 \\
& & \checkmark & 77.37 & 81.94 & 65.62 \\
\checkmark & \checkmark & \checkmark & 77.03 & 81.63 & 65.75 \\
\hline
\multicolumn{6}{l}{(f) Selection of Top-$k$ Queries} \\
\hline
\multicolumn{3}{c}{10 \textbf{(Ours)}} & \textbf{76.99} & \textbf{81.93} & \textbf{64.85} \\
\multicolumn{3}{c}{50} & 75.78 & 80.51 & 63.73 \\
\multicolumn{3}{c}{100} & 74.06 & 78.52 & 62.09 \\
\multicolumn{3}{c}{300} & 75.91 & 80.27 & 64.27 \\
\bottomrule
\end{tabular}}
\label{tab:ablation_study}
\end{table}

\noindent \textbf{Importance of Main Components.}
Our framework introduces two novel components beyond the baseline: FRM and RFRM. 
Table~\ref{tab:ablation_study}(a) shows that each component individually contributes to performance improvement. When combined, FRM and RFRM provide complementary benefits, resulting in a further increase in overall detection accuracy.

\noindent \textbf{Impact of Frequency Cues.
}The proposed FRM leverages both high- and low-frequency information to enrich image representations. 
To isolate its effect, we conduct experiments without incorporating RFRM, as reported in Table~\ref{tab:ablation_study}(b). 
The results indicate that both frequency components individually contribute to performance gains, while their combination yields the most significant improvement, highlighting their complementary nature.

\noindent \textbf{Importance of the Filter Generator Block.
}We assess the contribution of the Filter Generator Block (FGB), which dynamically decouples high- and low-frequency components within the FRM. 
As shown in Table~\ref{tab:ablation_study}(c), we compare dynamic filtering against a fixed configuration where $\alpha = \beta$ for both high- and low-pass filters. 
The results demonstrate that dynamic filtering not only eliminates the need for manually selecting a fixed threshold (thus reducing hyperparameter tuning) but also achieves the highest overall performance, underscoring its effectiveness.

\noindent \textbf{Importance of the Guidance Map.
}To analyze the role of the guidance map in RFRM, we decompose the module into two components: the projection layer and the guidance-based focus enhancement. 
As shown in Table~\ref{tab:ablation_study}(d), removing the guidance map leads to a substantial drop in performance, underscoring its critical role in directing attention toward regions of interest within the feature space.

\noindent \textbf{Selection of Feature Map Level.
}We investigate the impact of applying RFRM at different feature map resolutions. 
As shown in Table~\ref{tab:ablation_study}(e), applying RFRM to the highest-resolution feature map yields the best overall performance. 
In contrast, sharing RFRM across all three scales not only incurs substantial computational overhead but also leads to a decrease in accuracy.

\noindent \textbf{Selection of Top-k Queries.}
We examine the effect of various top-$k$ values when selecting queries for constructing the guidance map. As shown in Table~\ref{tab:ablation_study}(f), on the VOC dataset, characterized by a lower object-to-image ratio, a top-$k$ value of 10 is sufficient. Increasing $k$ beyond this threshold introduces noisy queries, which significantly degrades performance.

\subsection{Efficiency Analysis and Trade-off}

\begin{table}[t]
\centering
\caption{Efficiency analysis on multiple enhancers and detectors}
\renewcommand{\arraystretch}{1.0}
\begin{tabular}{@{}ccccc@{}}
\toprule
Model   & Tr. Time & Memory & FLOPs   & FPS  \\
\midrule
DINO    & 3h37m         & 22GB   & 233.09G & 26.6 \\
+ CPA     & 7h54m         & 110GB  & 389.88G & 10.1 \\
\midrule
RT-DETR  & 5h35m         & 2.9GB  & 61.12G  & 54.1 \\
+ MGDIP   & 35h13m        & 9.6GB  & 116.45G & 8.1  \\
+ CPA     & 15h47m        & 64GB   & 217.91G & 13.2 \\
\midrule
FR-DETR & 6h27m         & 3.8GB  & 96.04G  & 31.9 \\
\bottomrule
\end{tabular}
\label{tab:effi}
\end{table}

We further evaluate the computational efficiency of existing enhancement-based methods, as summarized in Table~\ref{tab:effi}. 
The comparison is conducted on two detectors: RT-DETR, trained for 48 epochs on a single A100 GPU, and DINO, trained for 36 epochs on two A100 GPUs. 
As observed, incorporating enhancers substantially increases FLOPs, reduces FPS, and requires significantly more GPU memory and training time for both detectors, while the performance gains remain marginal and in the case of DINO, even worse than the baseline. In comparison, FR-DETR achieves a significant improvement over the baseline RT-DETR and outperforms the accuracy of these heavy cascades, yet is drastically more efficient. Specifically, our method is \textbf{3x faster than MGDIP}, \textbf{2x faster than CPA}, and requires \textbf{17x less GPU memory} than CPA.

Finally, our detector-centric design is orthogonal to image-level enhancement. This modularity allows FR-DETR to be seamlessly cascaded with not only existing but also more advanced enhancers developed in the future. Consequently, for applications prioritizing maximum accuracy over latency, our framework serves as a robust foundation that can scale with advancements in low-level vision to further push the performance envelope.

\section{Conclusion}
We addressed limitations of enhancer-based pipelines for object detection under adverse weather, showing they introduce redundant computation and fail to focus on regions crucial for detection, resulting in minimal accuracy gains. To address this, we proposed FR-DETR, a detector-centric framework that performs feature-level refinement rather than image-level enhancement. Our approach integrates two components: (I) the Frequency Refinement Module, leveraging frequency-domain cues to improve foreground--background separation, and (II) the Recurrent Focus Refinement Module, which iteratively guides refinement toward regions of interest using coarse predictions. Extensive experiments demonstrate that FR-DETR achieves superior detection accuracy while being more computationally efficient than enhancer-based methods, enabling real-time deployment in adverse conditions.

\clearpage

\appendices

\section{Related Work}

\label{sec:rel_works_appendix}

\subsection{Object Detection in Adverse Conditions}

Classical detectors trained directly on degraded data (e.g., \cite{hendrycks2019robustness}) reveal substantial performance drops under occlusion, noise, and low contrast, indicating that learning robust features in adverse weather is challenging. 
A line of works mitigates degradations via pretrained restoration modules as preprocessing~\cite{histoformer,dehaze_rel}; however, despite improving perceptual quality, such restorers can hurt downstream detection~\cite{Gupta_2024_WACV}. 
Multi-task approaches~\cite{dsnet} attempt to jointly optimize restoration and detection, yet the objectives can conflict and hinder convergence or transfer, as discussed in~\cite{reforde}. 
Domain adaptation methods~\cite{dayolo,msdayolo} seek to bridge the gap between clean and degraded domains but face large appearance shifts and limited target supervision.
More recently, end-to-end enhancer-detector frameworks~\cite{iayolo,gdip,cpa} optimize both modules under a detection loss. 
These pipelines are typically benchmarked with YOLOv3~\cite{yolov3}; when paired with stronger modern detectors such as DINO and RT-DETR~\cite{dino,RTDETR}, they yield limited gains relative to their nontrivial computational overhead. 
Moreover, most image-level enhancers apply global adjustments rather than explicitly prioritizing object-centric regions that benefit detection.

\subsection{Frequency Domain Learning}

Frequency-domain representations have long been used in signal processing and are increasingly adopted in computer vision to disentangle foreground from background under low-contrast or cluttered scenes (e.g., camouflaged object detection/segmentation). 
Zhong et al.~\cite{Freq_camo_CVPR} draw inspiration from predator vision and introduce frequency-aware modules and losses to exploit frequency cues. 
Cong et al.~\cite{FPN_camo} employ octave convolution to factorize features into low- and high-frequency components, facilitating coarse-to-fine localization. 
Sun et al.~\cite{FSEL_eccv} jointly leverage global frequency information and local spatial features to obtain richer representations for camouflaged objects. 
Shi et al.~\cite{HFP} emphasize high-frequency signals to better delineate small objects, while Yan et al.~\cite{wavelet_sali} use wavelet-based decompositions to extract and fuse low-/high-frequency responses with transformer queries.

\subsection{Transformer-based Object Detection}
Transformer-based detectors have recently emerged as state-of-the-art, gradually replacing traditional CNN-based architectures. 
DETR~\cite{detr} introduced an end-to-end detection paradigm that eliminates hand-crafted components and post-processing (e.g., NMS) by predicting all objects in parallel using bipartite matching. 
However, DETR suffers from slow convergence and high computational cost. 
To address this, Deformable DETR~\cite{def_detr} replaces global attention with multi-scale deformable attention and introduces iterative bounding-box refinement, significantly accelerating training. 
DN-DETR~\cite{dn_detr} mitigates the instability of bipartite matching through a denoising strategy, while DINO~\cite{dino} further improves convergence and robustness via contrastive denoising training. 
Other works explore the limitations of one-to-one matching: H-DETR~\cite{hdetr} adds a one-to-many matching branch during training to enhance supervision. 
In parallel, lightweight DETR variants such as RT-DETR~\cite{RTDETR} have been proposed to achieve real-time performance while maintaining competitive accuracy, making them suitable for deployment in resource-constrained or latency-sensitive scenarios.

\section{Detailed Formulation of Mask Generation}
\label{sec:appendix_mask}

In the main paper, we introduced the Filter Generator Block (FGB), which predicts normalized cutoff parameters $(\alpha, \beta) \in [0,1]^2$ for the vertical (height) and horizontal (width) frequency axes. Here we provide the explicit derivation of the soft and hard masks used together with a Straight-Through Estimator (STE).

Let $H$ and $W$ denote the height and width of the feature map, respectively. We index discrete frequencies by $u \in \{0,\dots, H-1\}$ and $v \in \{0,\dots, W-1\}$, assuming an unshifted 2D DCT so that the DC component lies at $(u,v) = (0,0)$ in the top-left corner of the spectrum.

To obtain differentiable masks, we first define the probability that a coordinate belongs to the high-frequency region independently along each axis using a scaled sigmoid with sharpness $s$:
\begin{align}
    P_u(u) &= \sigma\bigl(s \, (u - \alpha H)\bigr), \\
    P_v(v) &= \sigma\bigl(s \, (v - \beta W)\bigr).
\end{align}
Here, $P_u(u) \approx 1$ when $u \gg \alpha H$ (high frequency) and $P_u(u) \approx 0$ when $u \ll \alpha H$ (low frequency), with a smooth transition near the cutoff.

\subsection{High-Pass Filter Construction}

The high-pass filter is designed to remove the DC and low-frequency components located in the top-left corner of the spectrum. The filter should be active (mask value 1) if \emph{either} the vertical or the horizontal frequency is high.

\paragraph{Hard mask (forward pass).}
We implement this with a logical OR:
\begin{equation}
    M_{\text{high}}^{\text{hard}}(u,v) =
    \begin{cases}
        1 & \text{if } u \ge \alpha H \ \lor\ v \ge \beta W, \\
        0 & \text{otherwise.}
    \end{cases}
\end{equation}
Equivalently, the mask is zero only for coordinates with $u < \alpha H$ \emph{and} $v < \beta W$, i.e., a low-frequency rectangle in the top-left corner.

\paragraph{Soft mask (backward pass).}
We approximate the logical OR by the probabilistic sum:
\begin{equation}
    M_{\text{high}}^{\text{soft}}(u,v)
    = 1 - \bigl(1 - P_u(u)\bigr)\,\bigl(1 - P_v(v)\bigr),
\end{equation}
which equals the probability that at least one axis is in the high-frequency range.

\subsection{Low-Pass Filter Construction}

The low-pass filter suppresses extreme high-frequency components located in the bottom-right corner. The filter should be active (mask value 1) unless \emph{both} vertical and horizontal frequencies are high.

\paragraph{Hard mask (forward pass).}
We use the logical NAND (i.e., pass if either dimension is low):
\begin{equation}
    M_{\text{low}}^{\text{hard}}(u,v) =
    \begin{cases}
        1 & \text{if } u < \alpha H \ \lor\ v < \beta W, \\
        0 & \text{otherwise.}
    \end{cases}
\end{equation}
Thus, only coordinates with $u \ge \alpha H$ and $v \ge \beta W$ (the bottom-right high--high block) are zeroed out, yielding an L-shaped passband.

\paragraph{Soft mask (backward pass).}
We approximate this by subtracting the probability of the intersection (logical AND) from 1:
\begin{equation}
    M_{\text{low}}^{\text{soft}}(u,v)
    = 1 - \bigl(P_u(u)\, P_v(v)\bigr),
\end{equation}
which is the probability that not both axes are high simultaneously.

\subsection{Gradient Propagation via STE}

Finally, we combine hard and soft masks using a straight-through estimator:
\begin{equation}
    M(u,v) = \mathrm{sg}\bigl(M^{\text{hard}}(u,v) - M^{\text{soft}}(u,v)\bigr)
             + M^{\text{soft}}(u,v),
\end{equation}
where $\mathrm{sg}(\cdot)$ denotes the stop-gradient operator. In the forward pass, $M(u,v)$ evaluates to the exact binary mask $M^{\text{hard}}$, while during backpropagation the gradients with respect to $\alpha$ and $\beta$ are those of the smooth masks $M^{\text{soft}}$.

\section{Dataset Details}
We provide further details on the evaluation benchmarks used in the main text.
First, we describe the standard single-degradation benchmarks commonly adopted in prior literature~\cite{gdip,erup}, specifically focusing on foggy and low-light conditions.
However, given that these datasets often lack diversity or severity, we explicitly discuss the composition of our compiled Adverse Weathers benchmark. This dataset introduces a more rigorous evaluation setting, featuring high-intensity synthetic and real-world degradations across fog, rain, and snow scenarios.

\noindent \textbf{Foggy Conditions.}
Following prior works~\cite{cpa,iayolo,gdip,erup}, we evaluate object detection under foggy conditions on five classes: \emph{person}, \emph{car}, \emph{bus}, \emph{bicycle}, and \emph{motorcycle}. 
We first construct a clear-weather training set by selecting Pascal VOC~\cite{pascal-voc-2007,pascal-voc-2012} train/val images containing at least one of the above classes, resulting in 8{,}111 images (denoted as V\_n\_tv). 
To simulate fog, we apply the atmospheric scattering model~\cite{ASM} as implemented in~\cite{iayolo} to generate a foggy counterpart (V\_f\_tv). 
For testing, we sample 2{,}734 images from the Pascal VOC 2007 test set (V\_f\_ts) and apply the same fog synthesis. 
Additionally, we include the RTTS dataset~\cite{rtts}, a real-world foggy benchmark with 4{,}322 images, for evaluation.

\noindent \textbf{Low-Light Conditions.}
For low-light experiments, we sample Pascal VOC images containing at least one of ten categories: \emph{bicycle}, \emph{boat}, \emph{bottle}, \emph{bus}, \emph{car}, \emph{cat}, \emph{chair}, \emph{dog}, \emph{motorbike}, and \emph{person}. 
This yields 12{,}334 images for the clear set (V\_n\_tv). 
We generate a low-light version (V\_d\_tv) by applying a gamma correction transformation $f(x)=x^\gamma$, where $\gamma$ is sampled from $\mathcal{U}(1.5,5)$ and $x$ denotes the input pixel intensity. 
The corresponding test set (V\_d\_ts) contains 3{,}760 images. 
We also evaluate on ExDark~\cite{Exdark}, a real-world low-light dataset with 2{,}563 images.

\noindent \textbf{Multi-weather Conditions.}
To overcome the limitations of existing benchmarks regarding limited diversity and insufficient intensity, we adopt the \textbf{Adverse Weathers} dataset, introduced in \cite{MODE}. This benchmark encompasses fog, rain, and snow at high intensities and is constructed by aggregating and rigorously modifying samples from FoggyCityscapes~\cite{SDV18}, RainCityscapes~\cite{Hu_2019_CVPR}, RainyCityscapes~\cite{halder2019physics}, RIS~\cite{Li2019DerainBenchmark}, BDD100K~\cite{bdd100k}, and ACDC~\cite{SDV21}. Consistent with the design philosophy of \cite{MODE}, we strictly utilize outdoor road-scene datasets to ensure physical realism, avoiding the domain inconsistencies often found in synthetic augmentations of indoor-heavy datasets like Pascal VOC~\cite{iayolo,gdip,cpa}.

\noindent \textbf{Dataset Composition.}
Following the protocol in \cite{MODE}, the dataset consists of 15,000 training images and 3,000 evaluation images. To ensure standard evaluation and label consistency across diverse sources, the \textit{rider} and \textit{pedestrian} classes are merged into a single \textit{person} category, resulting in six final classes: \textit{person, car, truck, bus, motorcycle,} and \textit{bicycle}.
For baseline comparisons, we also utilize the defined Clean Reference Set (6,532 images), which contains the non-degraded counterparts of the training samples.

\noindent \textbf{Training Set Details.}
As detailed in \cite{MODE}, the training set maintains a balanced distribution (4,600 images per synthetic category plus real-world samples):
\begin{itemize}
    \item \textbf{Fog (4,600 images):} Comprises 1,500 images sampled from FoggyCityscapes~\cite{SDV18} at three distinct intensity levels ($\beta \in \{0.005, 0.01, 0.02\}$), supplemented by 50 images processed at two higher intensities ($\beta \in \{0.01, 0.02\}$).
    \item \textbf{Rain (4,600 images):} Aggregates multiple sources, including 262 distinct images from RainCityscapes~\cite{Hu_2019_CVPR} (each augmented at three random intensities), 2,760 images from RainyCityscapes~\cite{halder2019physics}, and 1,054 images from RIS~\cite{Li2019DerainBenchmark}.
    \item \textbf{Snow (4,600 images):} Features realistic snow scenes generated by blending BDD100K~\cite{bdd100k} images with Snow100K~\cite{liu2018desnownet} masks. This subset includes 1,000 "multi-weather" samples (snow masks overlaying fog) and 300 images enhanced with synthetic generation techniques~\cite{Gupta_2024_WACV}.
    \item \textbf{Real-World Augmentation:} The set includes 400 real-world samples sourced from the ACDC training set~\cite{SDV21}, covering fog, rain, and snow conditions.
\end{itemize}

\noindent \textbf{Validation Set Details.}
For evaluation, we employ the official validation split defined in \cite{MODE}, designed to test generalization under severe conditions:
\begin{itemize}
    \item \textbf{Fog (900 images):} Sourced from FoggyCityscapes at three fixed degradation levels ($\beta \in \{0.005, 0.01, 0.02\}$).
    \item \textbf{Rain (900 images):} Consists of 33 images from RainCityscapes (at three intensities), 440 from RainyCityscapes, and 361 from RIS.
    \item \textbf{Snow (900 images):} Created by blending Snow100K masks with 650 BDD100K images and 250 FoggyCityscapes images. High-intensity augmentation~\cite{Gupta_2024_WACV} is applied to a subset of 50 images.
    \item \textbf{Real-World Evaluation:} An additional 300 real-world images from the ACDC dataset are included to benchmark performance against authentic weather artifacts.
\end{itemize}

\section{Additional Results}
In this section, we report more detailed results for each weather condition on our Adverse Weathers dataset. Finally, we present an ablation study to demonstrate our framework's generalizability.

\begin{table*}[t]
\centering
\caption{Detailed results on specific weather conditions in Adverse Weathers. Best and second-best scores are bold and underlined.}
\renewcommand{\arraystretch}{1.0}
\begin{tabular}{@{}llccccccccc@{}}
\toprule
\multirow{2}{*}{Approach} & \multirow{2}{*}{Model} & \multicolumn{3}{c}{Fog} & \multicolumn{3}{c}{Rain} & \multicolumn{3}{c}{Snow} \\ 
\cmidrule(lr){3-5} \cmidrule(lr){6-8} \cmidrule(lr){9-11}
 & & AP & AP$_{50}$ & AP$_{75}$ & AP & AP$_{50}$ & AP$_{75}$ & AP & AP$_{50}$ & AP$_{75}$ \\ 
\midrule
Direct & RT-DETR \cite{RTDETR} & 9.67 & 21.20 & 8.50 & 7.95 & 20.21 & 5.11 & 7.98 & 17.76 & 6.60 \\
\midrule
\multirow{4}{*}{Preprocessing} & RT-DETR \cite{RTDETR} & 6.47 & 13.85 & 5.20 & 2.53 & 7.14 & 1.39 & 3.37 & 8.57 & 2.41 \\
& Histoformer \cite{histoformer} & 6.68 & 14.21 & 5.37 & 2.67 & 7.44 & 1.58 & 5.63 & 12.61 & 4.62\\
& DCPT \cite{dcpt} & 6.49 & 13.81 & 5.40 & 2.47 & 6.93 & 1.48 & 4.61 & 10.72 & 3.44\\
& MoceIR \cite{moceir} & 6.37 & 13.79 & 5.19 & 2.42 & 6.75 & 1.38 & 4.95 & 11.09 & 3.93\\
\midrule
\multirow{1}{*}{Union} & DENet \cite{denet}  & 9.56 & 21.46 & 8.07 & 7.92 & 19.64 & 5.32 & 8.14 & 18.48 & 6.66\\
\midrule
Our & \textbf{FR-DETR} & \textbf{13.04}& \textbf{26.56}& \textbf{11.54} & \textbf{10.77} & \textbf{24.01} & \textbf{8.54} & \textbf{10.44} & \textbf{21.92} & \textbf{8.99}\\ 
\bottomrule
\end{tabular}
\label{tab:weather_spec}
\end{table*}

\noindent \textbf{Specific Degradations in Adverse Weather.} We present a detailed breakdown of results for three specific degradation types (fog, rain, and snow) within the Adverse Weather dataset. As shown in Table \ref{tab:weather_spec}, FR-DETR consistently outperforms competing baselines across all three metrics and weather conditions. We achieve a particularly significant margin over RT-DETR, with gains of 3.37 and 5.36 in $\text{AP}$ and $\text{AP}_{50}$ for foggy conditions, respectively. This demonstrates our method's ability to ensure robustness against adverse weather without the high computational overhead of current enhancer-based approaches.

\begin{table}[ht]
\centering
\caption{Generalization of FR-DETR on ResNet-34VD backbone}
\renewcommand{\arraystretch}{1.0}
\begin{tabular}{@{}cccc@{}}
\toprule
Backbone   & AP & AP$_{50}$ & AP$_{75}$  \\
\midrule
RT-DETR & 11.60 & 25.01 & 10.22 \\
FR-DETR & \textbf{12.71} & \textbf{25.53} & \textbf{11.11} \\
\bottomrule
\end{tabular}
\label{tab:scaled}
\end{table}

\noindent \textbf{Generalizability and Backbone Scaling.} To assess scalability, we apply FR-DETR to the larger ResNet-34VD backbone. As shown in Table \ref{tab:scaled}, our framework generalizes well, consistently outperforming the RT-DETR baseline across all metrics. However, we observe that the performance gap is narrower compared to the ResNet-18VD experiments. This aligns with our findings on the milder V\_f\_ts dataset in the main paper: larger backbones (or easier datasets) naturally yield more robust features, leaving less room for improvement. Consequently, this suggests that FR-DETR is most critical in challenging scenarios, specifically when using lightweight backbones or facing severe degradations, effectively bridging the gap between efficient models and heavy-duty performance.

\section{Visualization}

\subsection{Feature Visualization}

\begin{figure}[ht]
\begin{center} 
   \includegraphics[width=1\linewidth]{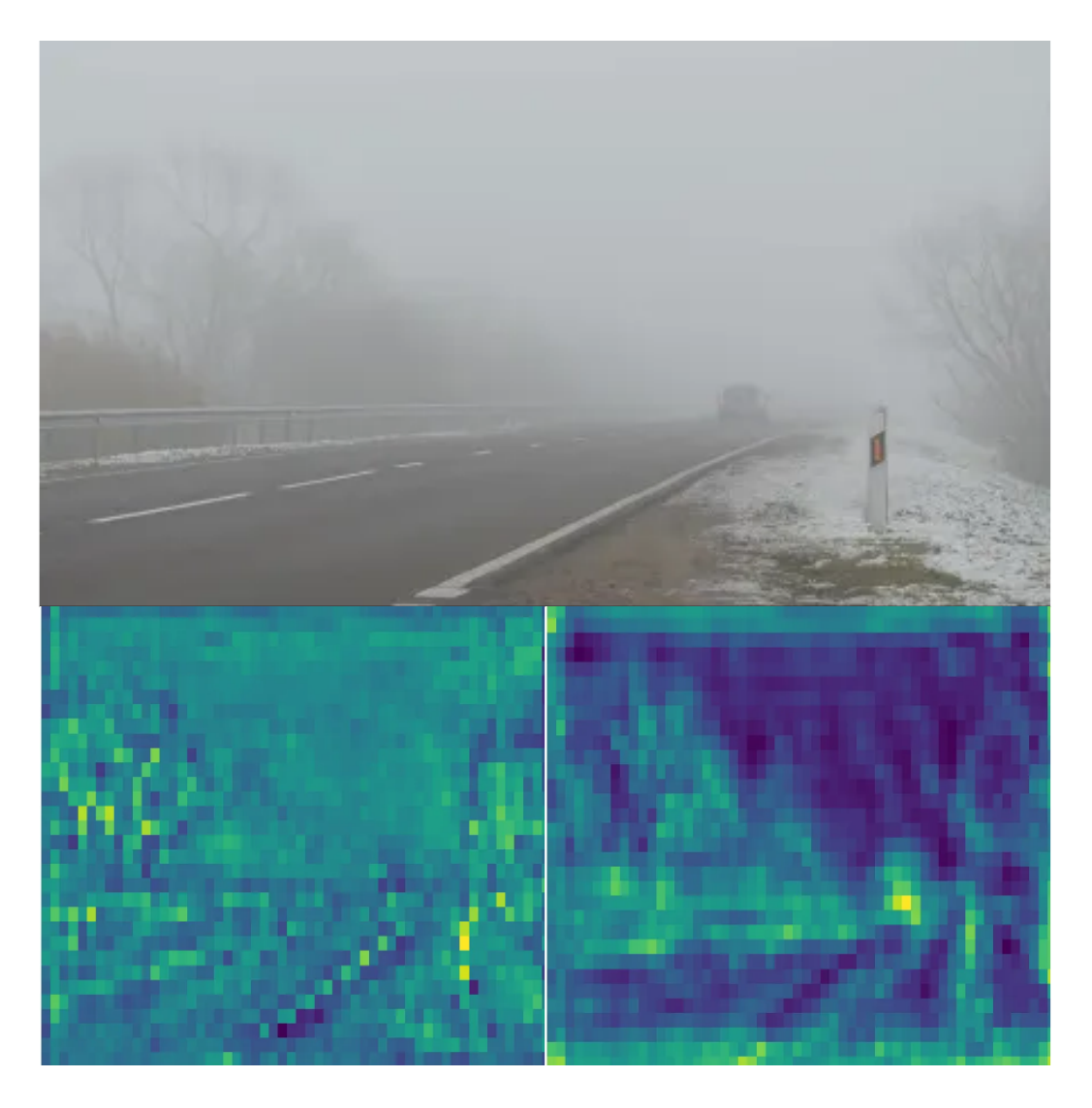}
\end{center}
   \caption{Top image is the input image while bottom left and bottom right ones are feature map before and after FRM.}
\label{fig:FRM_vis}
\end{figure}

\begin{figure}[ht]
\begin{center} 
   \includegraphics[width=1\linewidth]{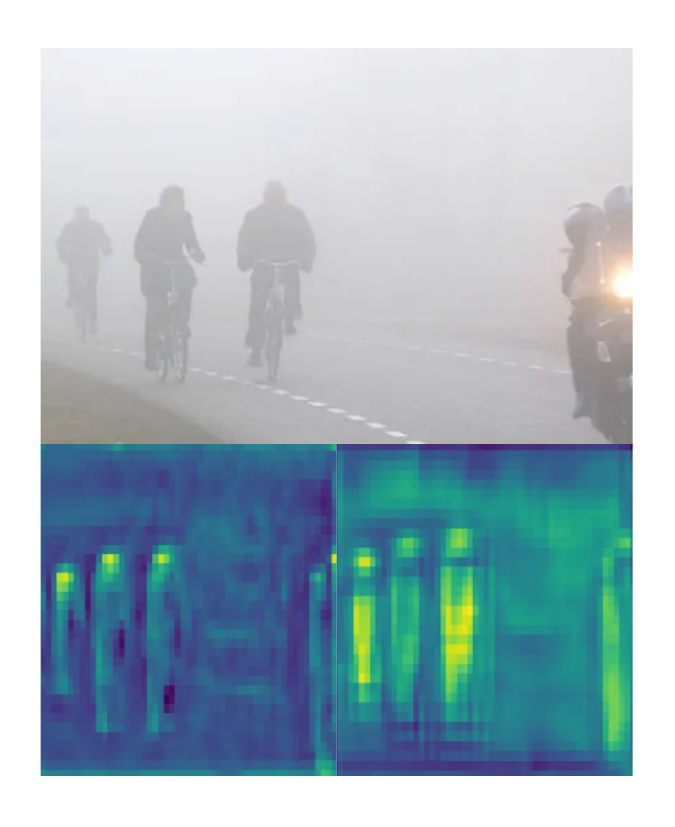}
\end{center}
   \caption{Top image is the input image while bottom left and bottom right ones are feature map before and after RFRM.}
\label{fig:RFRM_vis}
\end{figure}

We visualize feature maps before and after each proposed component in Fig.~\ref{fig:FRM_vis} and Fig.~\ref{fig:RFRM_vis}. 
In Fig.~\ref{fig:FRM_vis}, distant vehicles are heavily attenuated by fog; prior to FRM, activations are diffuse and dominated by the background veil, whereas after FRM they concentrate around the far car, indicating improved foreground-background separability. 
Similarly, in Fig.~\ref{fig:RFRM_vis}, the initial responses predominantly highlight the pedestrian's head; following RFRM, activations extend to the full body and the bicycle, better capturing object extent and boundaries. 
These qualitative results align with our design goals, frequency-aware feature shaping (FRM) and prediction-guided focusing (RFRM), and corroborate the quantitative improvements.

\subsection{Results Visualization.}

\begin{figure*}[ht]
\begin{center} 
   \includegraphics[width=1\linewidth]{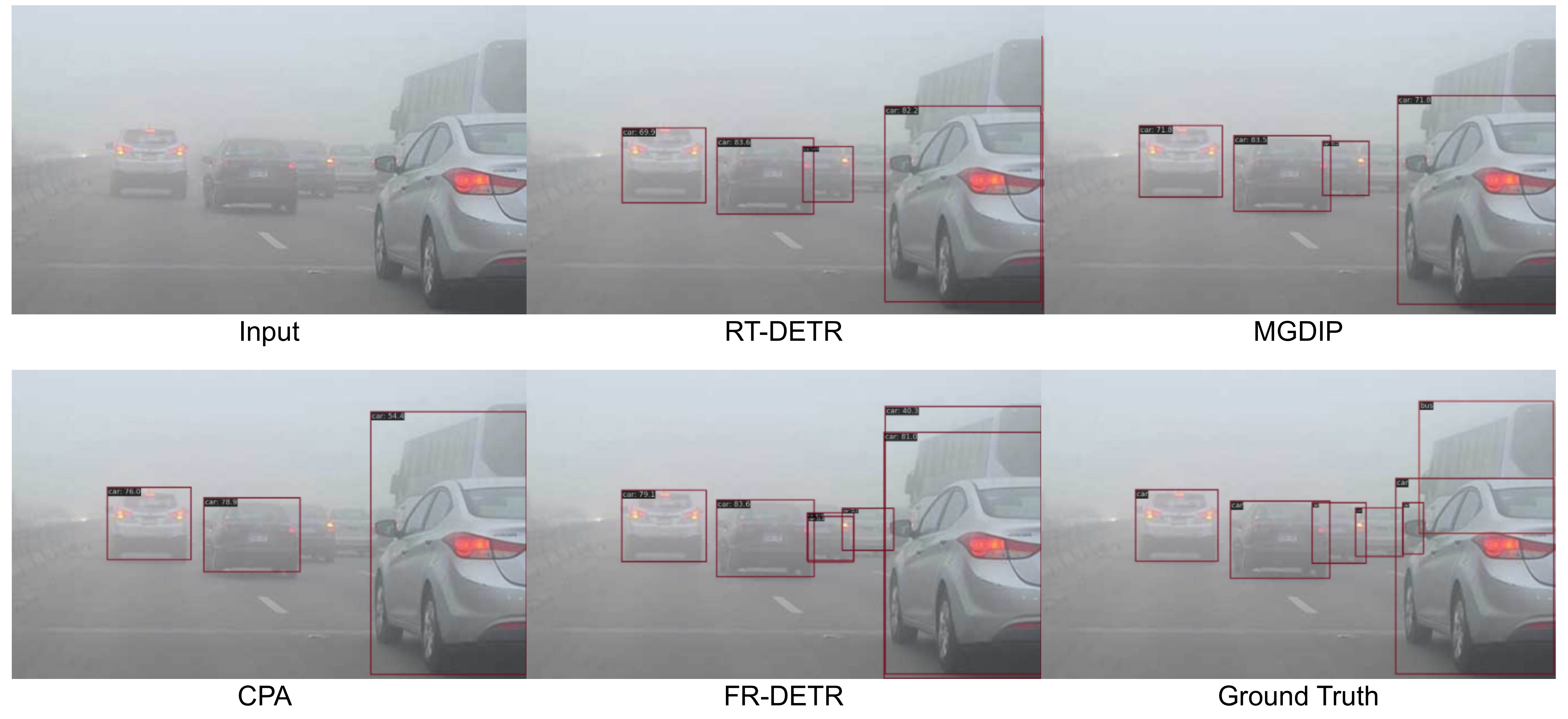}
\end{center}
   \caption{Visualization of detection results trained on a 5-class model in foggy conditions.}
\label{fig:vis_fog5}
\end{figure*}

\begin{figure*}[ht]
\begin{center} 
   \includegraphics[width=1\linewidth]{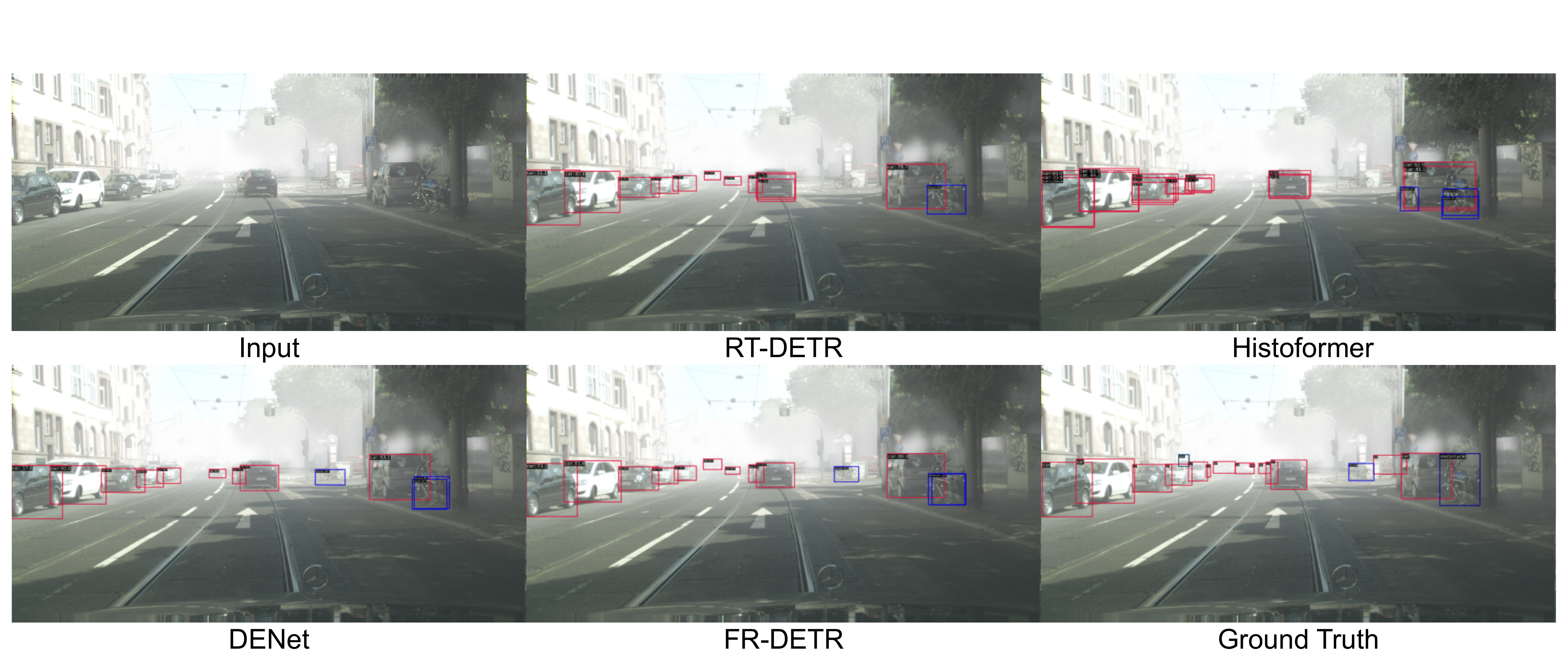}
\end{center}
   \caption{Visualization of detection results trained on a 6-class model in foggy conditions.}
\label{fig:vis_fog6}
\end{figure*}

\begin{figure*}[ht]
\begin{center} 
   \includegraphics[width=1\linewidth]{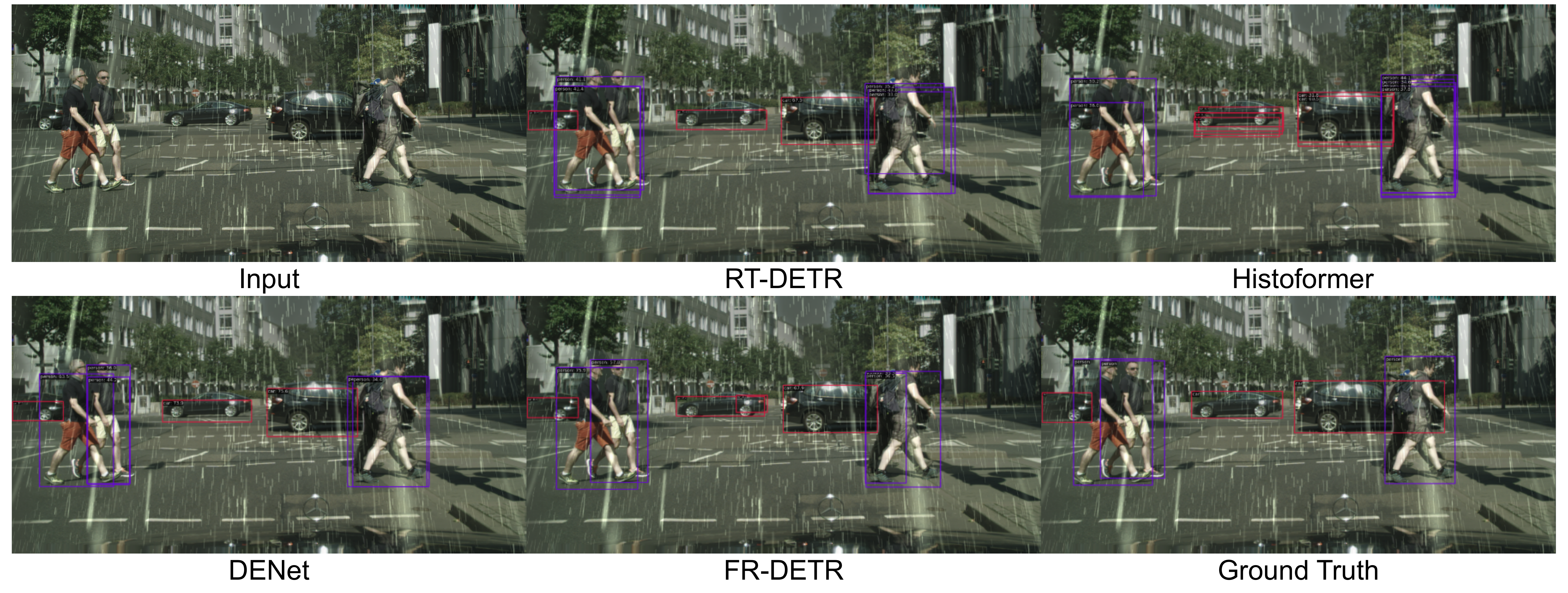}
\end{center}
   \caption{Visualization of detection results trained on a 6-class model in rainy conditions.}
\label{fig:vis_rain6}
\end{figure*}

\begin{figure*}[ht]
\begin{center} 
   \includegraphics[width=1\linewidth]{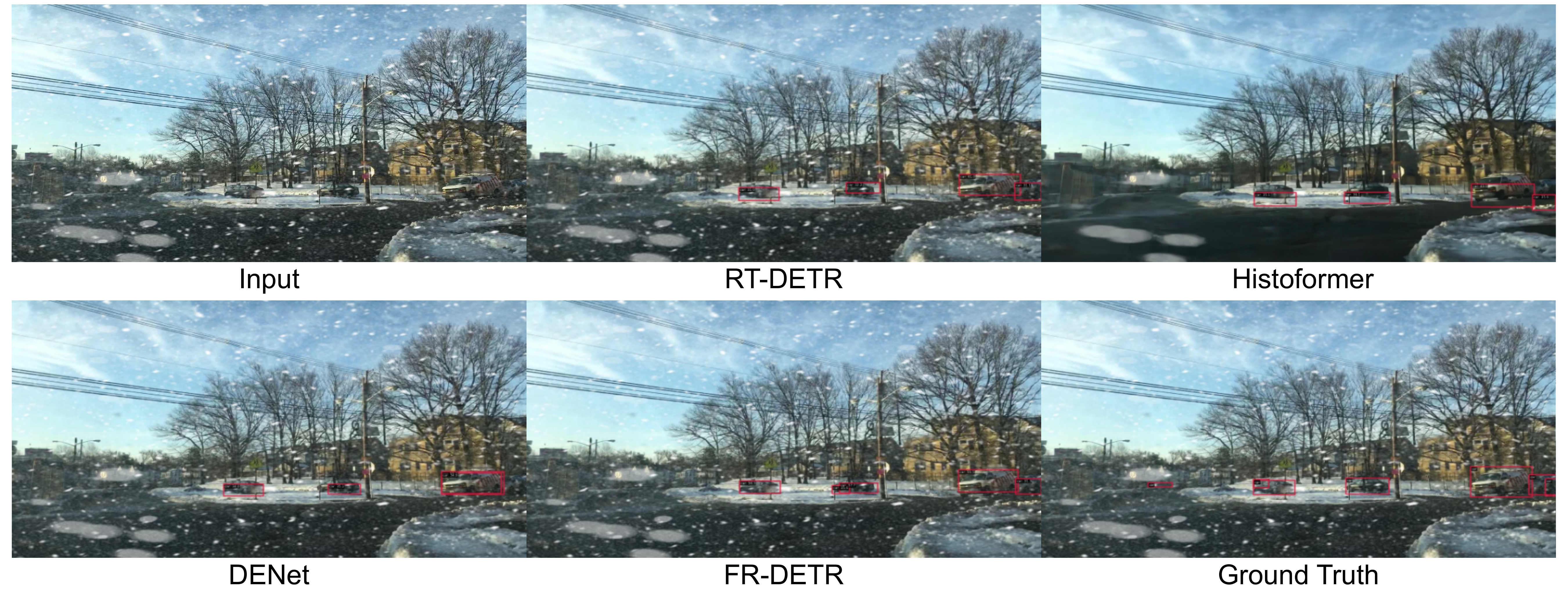}
\end{center}
   \caption{Visualization of detection results trained on a 6-class model in snowy conditions.}
\label{fig:vis_snow6}
\end{figure*}

Here we provide visualizations of the detection results of our proposed method FR-DETR compared to other baselines. Figure \ref{fig:vis_fog5} shows the detection results when trained under foggy conditions using a 5-class model. In addition, we also include visualizations of our method trained on the Adverse Weathers dataset under three different degradations (foggy, rainy, and snowy) in Figure \ref{fig:vis_fog6}, Figure \ref{fig:vis_rain6}, and Figure \ref{fig:vis_snow6}, respectively.

\clearpage
{
    \balance
    \bibliographystyle{IEEEbib}
    \bibliography{icme2026references}
}

\end{document}